\documentclass[10pt,twocolumn,letterpaper]{article}

\usepackage{cvpr}              
\usepackage{kotex}
\usepackage{makecell}
\usepackage{siunitx}
\usepackage{indentfirst}
\usepackage[accsupp]{axessibility}
\definecolor{cvprblue}{rgb}{0.21,0.49,0.74}
\usepackage[pagebackref,breaklinks,colorlinks,citecolor=cvprblue]{hyperref}

\def\paperID{00049} 
\def\confName{CVPR}
\def\confYear{2024}

\title{VisTA-SR: Improving the Accuracy and Resolution of Low-Cost Thermal Imaging Cameras for Agriculture}

\author{
Heesup Yun, Sassoum Lo, Christine H. Diepenbrock, Brian N. Bailey, J. Mason Earles\\
University of California, Davis\\
1 Shields Ave, Davis, CA 95616\\
{\tt\small \{hspyun, ssslo, chdiepenbrock, bnbailey, jmearles\}@ucdavis.edu}
}

\begin{document}
\maketitle
\begin{abstract}
\indent Thermal cameras are an important tool for agricultural research because they allow for non-invasive measurement of plant temperature, which relates to important photochemical, hydraulic, and agronomic traits. Utilizing low-cost thermal cameras can lower the barrier to introducing thermal imaging in agricultural research and production. This paper presents an approach to improve the temperature accuracy and image quality of low-cost thermal imaging cameras for agricultural applications. Leveraging advancements in computer vision techniques, particularly deep learning networks, we propose a method, called $\textbf{VisTA-SR}$ ($\textbf{Vis}$ual \& $\textbf{T}$hermal $\textbf{A}$lignment and $\textbf{S}$uper-$\textbf{R}$esolution Enhancement) that combines RGB and thermal images to enhance the capabilities of low-resolution thermal cameras. The research includes calibration and validation of temperature measurements, acquisition of paired image datasets, and the development of a deep learning network tailored for agricultural thermal imaging. Our study addresses the challenges of image enhancement in the agricultural domain and explores the potential of low-cost thermal cameras to replace high-resolution industrial cameras. Experimental results demonstrate the effectiveness of our approach in enhancing temperature accuracy and image sharpness, paving the way for more accessible and efficient thermal imaging solutions in agriculture.
\end{abstract}    
\section{Introduction}
\label{sec:intro}
\begin{figure*}[h]
    \centering
    \includegraphics[width=1.0\linewidth]{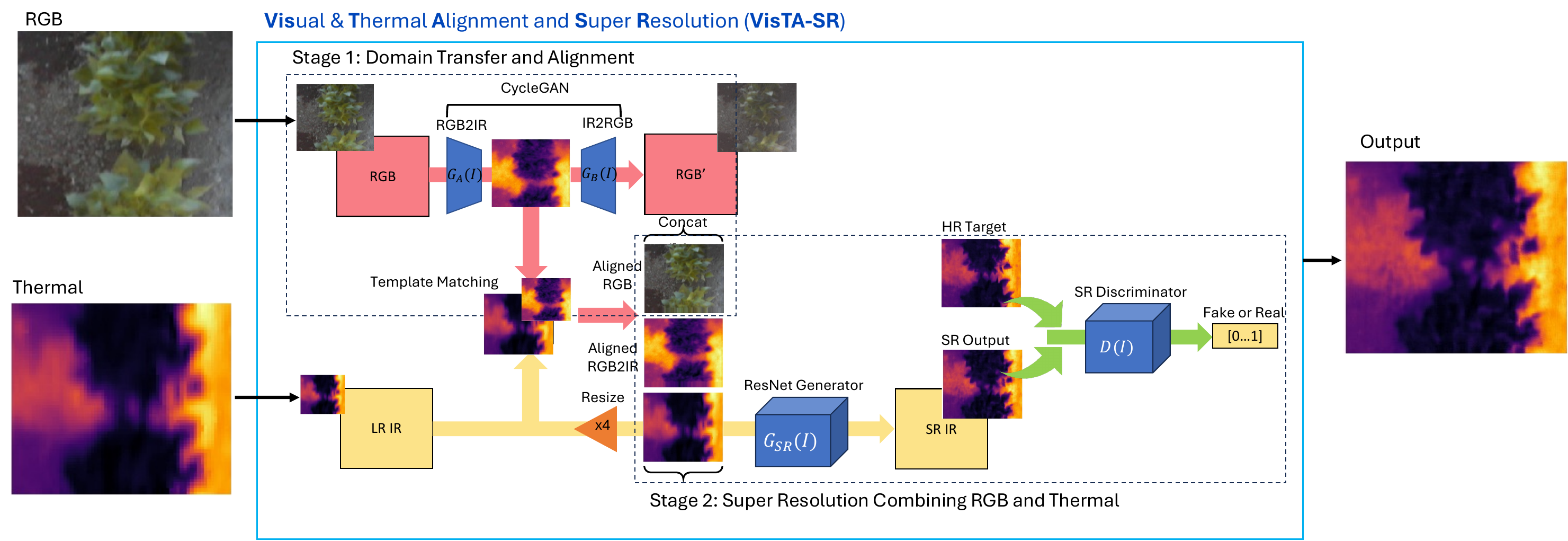}
     \caption{Structure of the proposed VisTA-SR network. The network has two main stages: the Image Alignment and the Super-Resolution Network. The Image Alignment aligns the RGB and thermal images, while the Super-Resolution Network enhances the resolution of the thermal image.}
     \label{fig:front}
\end{figure*}

Agricultural research often uses crop temperature measurement to detect abnormal plant characteristics, calculate crop water stress indices, or model complex biophysical interactions. Since various methods have been attempted to measure crop temperature, thermal imaging cameras are widely used because they can quickly measure the temperature at many points in the image~\cite{khorsandiPlantTemperaturebasedIndices2018}. Also, thermal imaging can quickly and non-invasively measure crop temperature compared to other temperature measurement devices. These benefits can help identify areas where crops are experiencing disease or stress, allowing for timely intervention.

Previous studies using thermal cameras in agriculture have utilized high-resolution industrial-grade thermal cameras~\cite{gonzalez-dugoUsingHighResolution2013}. However, these cameras are very expensive, often costing over \$10,000 which limits their accessibility. This low accessibility can restrict the widespread deployment of thermal cameras in agriculture, especially for researchers who cannot afford costly sensors.

An alternative approach is to use low-cost sensors. Recent developments in thermal image sensors and image processing technologies have made various affordable consumer-grade thermal cameras available. These thermal cameras have the advantages of being relatively lightweight and easy to operate. Therefore, there have been attempts to use low-cost thermal cameras in agriculture~\cite{gimenez-gallegoIntelligentThermalImagebased2021,gimenez-gallegoAutomaticCropCanopy2023,blaya-rosFeasibilityLowCostThermal2020}. For example, Bhandari~\cite{bhandariUseInfraredThermal2016} obtained an image mask from visible light images and applied it to thermal images to measure wheat canopy temperature and estimate water stress. Another study used a low-cost thermal camera to calculate crop canopy temperature automatically~\cite{gimenez-gallegoAutomaticCropCanopy2023}. However, these low-cost thermal cameras have not been able to completely replace high-resolution industrial thermal cameras due to their lower pixel count and resolution.

Thermal camera resolution has a significant impact on the capability and accuracy of agricultural research. For example, low-resolution thermal cameras may only be able to recognize crops at the plant-level rather than organ-level, making it challenging to observe temperature differences between leaves, stems, flowers, and fruits, for instance. This limited feature resolution will limit the temperature measurement capability at various phenological stages, which is essential for developing precise crop biophysical models. Therefore, improving the quality of low-cost thermal images can increase the feasibility of using low-cost thermal cameras in agriculture.

Enhancing the resolution of low-resolution thermal images is a challenging task. It is an ill-posed problem, as multiple high-resolution ground truths can exist for a single low-resolution image. Nevertheless, various computer vision and machine learning techniques have been proposed to overcome the challenge. Particularly with the recent advancements in deep learning, there have been many reported cases of upsampling low-resolution images to high-resolution. Some researchers have used ResNet and GANs to perform image super-resolution~\cite{ledigPhotoRealisticSingleImage2017}. Others have combined multiple low-resolution images to create a single high-resolution image~\cite{rivadeneiraMultiImageSuperResolutionThermal2022}. Some have also used multimodal data to improve the resolution of the data~\cite{guptaUnalignedGuidedThermal2022, almasriRGBGuidedThermal2018, chenColorGuidedThermal2016}. However, research on improving the quality of thermal images in the agricultural domain has been limited. Applying these techniques to agricultural thermal images could potentially improve the image quality of low-resolution thermal cameras, allowing them to replace high-resolution thermal cameras.

Therefore, this paper studies how computer vision techniques can improve the image quality of low-resolution thermal cameras for agricultural applications. We propose a deep learning network that leverages complementary information from RGB and thermal image domains for both image alignment and super-resolution enhancement.

The specific contributions of this paper are as follows:

\begin{itemize}
    \item Calibration and validation of the temperature measurement of a low-cost thermal camera in the agricultural domain
    \item Acquisition of a paired low-resolution thermal camera image dataset, as well as RGB and high-resolution thermal camera data in the agricultural domain
    \item Proposal of an integrated image alignment and super-resolution deep learning algorithm to improve the image quality of low-resolution thermal cameras by combining RGB and thermal images
\end{itemize}
\section{Related Work}
\label{sec:formatting}

\subsection{Traditional Image Enhancement}
Before the advent of deep learning-based image sharpening approaches, filter-based techniques were used to enhance image quality, including fundamental Gaussian kernels and image sharpening kernels such as Bilinear filtering~\cite{smithBilinearInterpolationDigital1981}, Bilateral filtering~\cite{tomasiBilateralFilteringGray1998}, and Lanczos filtering~\cite{duchonLanczosFilteringOne1979}. Despite their ability to reduce image noise and enhance object edges, these approaches have been criticized for introducing artificial noises not present in the original image or producing unsatisfactory sharpness.

\subsection{Deep Learning Based Super Resolution}
Recently, there have been attempts to improve sharpness using deep learning. These attempts include making super-resolution images from low-resolution images using ResNet and GANs, resulting in various developed methods~\cite{ledigPhotoRealisticSingleImage2017,wangESRGANEnhancedSuperResolution2018, chenActivatingMorePixels2023}. These methods have shown the ability to restore low-resolution images with higher quality compared to traditional filter-based algorithms. However, they still face challenges in overcoming the ill-posed problem of creating shapes that do not exist in the original image.

\subsection{Multi-Image or Multi-Modal Super Resolution}
To address the ill-posed problem in super-resolution methods, attempts have been made to create a single high-resolution image using various low-resolution or complementary information. For example, one approach is to utilize the high-resolution panchromatic channel of satellite imagery to enhance the sharpness of lower-resolution channels~\cite{dadrassjavanReviewImageFusion2021}. Another approach is to combine information from multiple frames to improve the sharpness of thermal images~\cite{cascaranoSuperResolutionThermalImages2020}. Additionally, multi-modal super-resolution techniques that combine RGB and thermal information have also been tried~\cite{guptaUnalignedGuidedThermal2022, almasriRGBGuidedThermal2018, chenColorGuidedThermal2016}.

\subsection{Use Cases of Thermal Cameras in Agriculture}
Most agricultural research studies have traditionally relied on high-resolution thermal cameras in their research. For example, Gonzalez-Dugo \etal~\cite{gonzalez-dugoUsingHighResolution2013} showed promising results assessing water stress within a commercial orchard using a high-resolution thermal camera, which costs more than \$20,000. Yan \etal~\cite{yanHighTemporalSpatial2023} recently employed a Pro SC TIR camera (640x512 resolution, \$17,250) to estimate evaporation, transpiration, and evapotranspiration, crucial parameters for understanding water dynamics in agricultural systems. However, these cameras can be prohibitively expensive, limiting their accessibility for many researchers and farmers.

In recent years, the emergence of low-cost thermal cameras has opened up new possibilities for agricultural applications. Several studies have explored the use of low-cost thermal cameras in agricultural research. García-Tejero \etal~\cite{garcia-tejeroAssessingCropWaterStatus2018} compared the performance of a low-cost FLIR One camera (80x60 resolution, \$400) with a high-end FLIR SC660 camera (640x480 resolution, \$20,000) for assessing crop water status. They found that the low-cost camera was able to provide valuable insights, demonstrating the potential for more affordable thermal imaging solutions. Similarly, Iseki \etal~\cite{isekiNewIndicatorLeaf2020} used a FLIR C2 camera (80x60 resolution, \$500) to estimate leaf stomatal conductance, a key indicator of plant water status. Parihar \etal~\cite{pariharApplicationInfraredThermography2021} utilized a FLIR E6 camera (240x180 resolution, \$2,000) for irrigation scheduling of horticultural plants, demonstrating its utility in optimizing water use. While low-cost thermal cameras offer an attractive alternative, their lower resolution and image quality than their high-end counterparts may limit their ability to provide the same level of detailed information. Additionally, the temperature accuracy of low-cost cameras in various environmental conditions and crop types needs further investigation. Nonetheless, the studies reviewed here highlight the potential of low-cost thermal cameras in agricultural research.
\section{Materials and Methods}
\subsection{Thermal Cameras}
In this study, three types of thermal cameras were utilized. Table \ref{tab:infrared_cameras} shows the specifications of the thermal cameras. The VarioCAM HD camera, known for its high spatial resolution and temperature accuracy, was primarily used to create a dataset for temperature accuracy validation. The VarioCAM HD images were collected using their proprietary software on the Windows Operating System. The FLIR Boson camera, with a resolution of 640x512, was employed to capture high-resolution thermal image data in the field. Positioned between high-end and consumer-grade thermal cameras in terms of price, the FLIR Boson camera offered a lightweight form factor and flexible video output interface for easy field image capture. FLIR Boson images were collected from the ROS-based system on Ubuntu PC. Lastly, the FLIR One Pro, a low-cost and low-resolution thermal camera, was used in this study. It has a thermal resolution of 160x120 and an RGB camera resolution of 1440x1080. FLIR One Pro image acquisition and storage were performed using a custom Swift-based app developed with the FLIR Mobile API on an iPhone.

\begin{table*}
    \centering
    \begin{tabular}{@{}lccc@{}}
    \toprule
    & VarioCam HD Head 800 & FLIR Boson & FLIR One Pro \\
    \midrule
    Spectral Range & 7.5 - 14 µm & 8 - 14 µm & 8 - 14 µm \\
    Detector Resolution & 1,024 × 768 & 640 × 512 & 160 × 120 \\
    Temperature Measuring Range & -40 - 2,000 °C & Non Radiometric & \makecell{-20 - 120 °C}\\
    Measurement Accuracy & ±1.5°C or ±1.5\% & Non Radiometric & ±3°C or ±5\% \\
    Temperature Sensitivity & 30mK & 40mK & 70 mK \\
    Frame Rate & 30 Hz \& 60 Hz & 9 Hz & 8.7 Hz \\
    Dimensions & 221 × 90 × 94 mm & 21 x 21 x 11 mm& 68 × 34 × 14 mm \\
    Weight & 1.15 kg & 21g & 36.5 g \\
    Price (Approx.) & \$20,000 & \$4,000 & \$400 \\
    \bottomrule
    \end{tabular}
    \caption{Specifications of thermal cameras used in this study}
    \label{tab:infrared_cameras}
\end{table*}

\subsection{Low Cost Thermal Camera Calibration}


Radiometric thermal cameras have a logarithmic relationship between the digital number and temperature~\cite{tattersallInfraredThermographyNoninvasive2016, minkinaInfraredThermographyErrors2009}. The parameters for converting the digital number to temperature are stored in the EXIF tag information of the FLIR radiometric JPEG images. These parameters, which are pre-calibrated values from the factory, are used to convert the digital numbers of the thermal imaging camera to temperatures using Equation \ref{eq:flir_one_pro}. Upon comparing the factory parameters of different thermal imaging cameras, it was observed that only the values of $R_1$ and $O$ differed, while the values of $R_2$, $B$, and $F$ remained constant. The parameter $B$ is derived from the Planck constant $h$ and Boltzmann constant $k_b$, and the parameter F value is 1. For the FLIR One Pro cameras, $R_2$ was fixed at 0.0125, and $R_1$ and $O$ are empirically calibrated depending on the individual camera. 

\begin{equation}
    \text{Temperature }(\SI{}{\celsius}) = \frac{B}{\ln(\frac{R_1}{R_2(DN+O)})+F} - 273.15
    \label{eq:flir_one_pro}
\end{equation}

However, the accuracy of these factory parameters cannot be fully trusted as the manufacturer does not fully guarantee the temperature accuracy of the low-cost thermal cameras. To ensure the accuracy of temperature measurements, it is necessary to recalibrate the parameters of the thermal imaging camera. Therefore, the optimization process focused on optimizing the values of $R_1$ and $O$. The optimization was performed using the Nelder-Mead method~\cite{nelderSimplexMethodFunction1965}, which is a widely used optimization algorithm, with a tolerance of $1e-6$. The optimization process was implemented using the `scipy.optimize.minimize' function in Python.

Experiments were conducted to verify the temperature accuracy of the FLIR One Pro thermal imaging camera. The surface temperature of a controlled water bath was measured using a thermocouple with a digital data logger. The thermocouple measured the temperature starting from $\SI{4.0}{\celsius}$, the initial temperature of the cold water, and reaching $\SI{100.0}{\celsius}$, the boiling point of water. The air temperature and relative humidity were maintained during the experiment at $\SI{24.0}{\celsius}$ and 40\%.

\begin{figure}[t]
    \centering
    \includegraphics[width=0.8\linewidth]{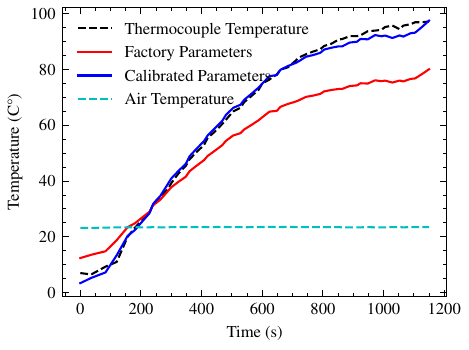}
     \caption{Comparison between thermocouple, factory, and calibrated temperature values in a time series}
     \label{fig:calibration_time}
\end{figure}

Figure \ref{fig:calibration_time} shows the temperatures calculated using the factory parameters. The results showed that the temperatures calculated using the factory parameters were higher than reference temperatures below $\SI{30}{\celsius}$. However, at temperatures near the boiling point of water, the measured temperatures were almost $\SI{20}{\celsius}$ lower than the actual temperatures. This indicates that the low-cost thermal camera's temperature values are inaccurate, especially at high temperatures.

\begin{figure}[t]
    \centering
    \includegraphics[width=0.8\linewidth]{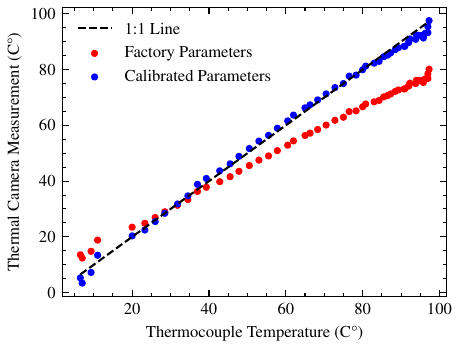}
     \caption{Comparison between factory and calibrated temperature values in a 1:1 plot}
    \label{fig:calibration_1on1}
\end{figure}

The original and optimized parameters are shown in Table \ref{tab:parameter-comparison}, and the temperatures calculated using the optimized new parameters are shown in Figure \ref{fig:calibration_time}. The temperatures calculated using the new parameters are more accurate than the results using the factory parameters, and they are almost identical to the temperatures measured by the thermocouple (Figure \ref{fig:calibration_1on1}).

\begin{table}[h]
    \centering
    \begin{tabular}{lrrrrr}
    \toprule
    & \textbf{$R_1$} & $B$ & $F$ & \textbf{$O$} & $R_2$ \\
    \midrule
    Factory & \textbf{18333.4} & 1435 & 1 & \textbf{-2284} & 0.0125 \\
    Optimized & \textbf{12755.4} & 1435 & 1 & \textbf{-6707} & 0.0125 \\
    \bottomrule
    \end{tabular}
    \caption{Comparison of factory and lab calibrated parameters}
    \label{tab:parameter-comparison}
\end{table}

\subsection{Matching Low-Resolution and High-Resolution Thermal Imaging Cameras}

As part of a more extensive set of breeding experiments, Cowpea (\textit{Vigna unguiculata L. Walp.}) and Common Bean (\textit{Phaseolus vulgaris}) images were collected from June to September 2022 in Davis, California, to obtain high-resolution and low-resolution thermal images in the field. To match the low-resolution and high-resolution thermal imaging datasets, camera calibration was performed to calculate each camera's intrinsic parameters, and camera extrinsics were also measured. The two cameras were installed at a height of approximately 1.5m from the ground, and the distance between the centers of the two camera lenses was 5cm.

However, the high-resolution and low-resolution thermal images were captured on different platforms at different frame rates, so matching the two datasets was challenging. Initially, SIFT~\cite{{loweDistinctiveImageFeatures2004}} feature extraction and matching were attempted, similar to the previous temperature accuracy test. However, the quality and the number of the extracted features in the images sometimes incorrectly estimate the homography between the low-resolution and high-resolution pair, which led to unstable matching and alignment results.

Since the field of view difference between the two images is only due to the scale difference based on the image resolution and the transitional offsets caused by the capture timings, template matching~\cite{sarvaiyaImageRegistrationTemplate2009} was performed to robustly match the images by setting the high-resolution image as the template image $T$ and calculating the Normalized Cross Correlation (NCC)~\cite{briechleTemplateMatchingUsing2001} between the template image and the low-resolution image $I$, finding the coordinates $x^*$ and $y^*$ where the NCC value was maximized, and resizing the template image to a predefined scale for this process.
 
\begin{equation}
R(x,y)=\sqrt{\frac{\sum_{x',y'}(T(x',y')-I(x+x',y+y'))^2}{\sum_{x',y'}T(x',y')^2\cdot\sum_{x',y'}I(x+x',y+y')^2}}
\end{equation}

\begin{equation}
(x^*, y^*) = \text{argmax}_{\substack{0 \leq x < M \\ 0 \leq y < N}} R(x,y) 
\end{equation}

\begin{figure*}
    \centering
    \includegraphics[width=1.0\linewidth]{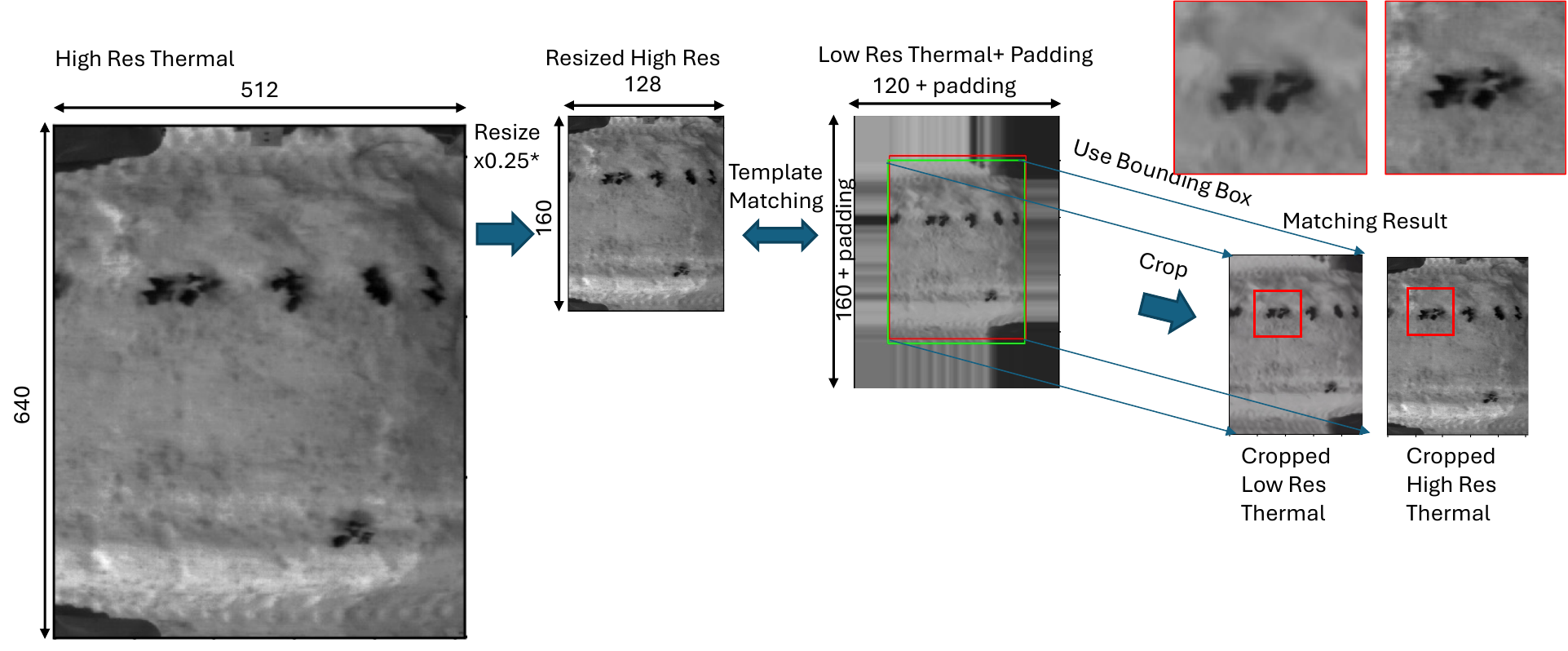}
     \caption{Matching and aligning process of low-resolution and high-resolution thermal images}
     \label{fig:template_matching_flow}
  \end{figure*}

The template matching was performed using Python OpenCV code. Figure \ref{fig:template_matching_flow} illustrates matching the low-resolution and high-resolution thermal imaging cameras. For cases where the NCC value was 0.75 or higher, the bounding box was calculated and limited to the area within the padding of the low-resolution thermal image coordinate system. Then, it was converted to the coordinate system before resizing the template image. Image cropping was performed using the original resolution of the template and background images. The FLIR One Pro also has an integrated RGB camera, allowing simultaneous acquisition of RGB images. Therefore, the RGB images were also cropped using the Template Matching results.

\subsection{Improving Image Resolution by Combining RGB and Thermal Imaging}

In this paper, complementary information from the RGB image's structural details and the thermal imaging camera's intensity information is utilized to enhance the resolution of the low-resolution thermal imaging camera. The RGB and thermal images obtained from the FLIR One Pro have the same field of view, but they are not perfectly pixel-aligned due to differences in camera lens position and video stream delays, which poses a challenge in combining the two modalities for resolution improvement. We tested deep-learning based image registration methods such as Spatial Transformer Networks~\cite{jaderbergSpatialTransformerNetworks2015} and Deformable Field-based approaches~\cite{zouReviewDeepLearningbased2022}. However, those methods tended to learn a shortcut existing in the dataset, which is a mean offset of the images rather than the differences between the input images, resulting in unstable experimental results.

\begin{figure}
    \centering
    \includegraphics[width=1.0\linewidth]{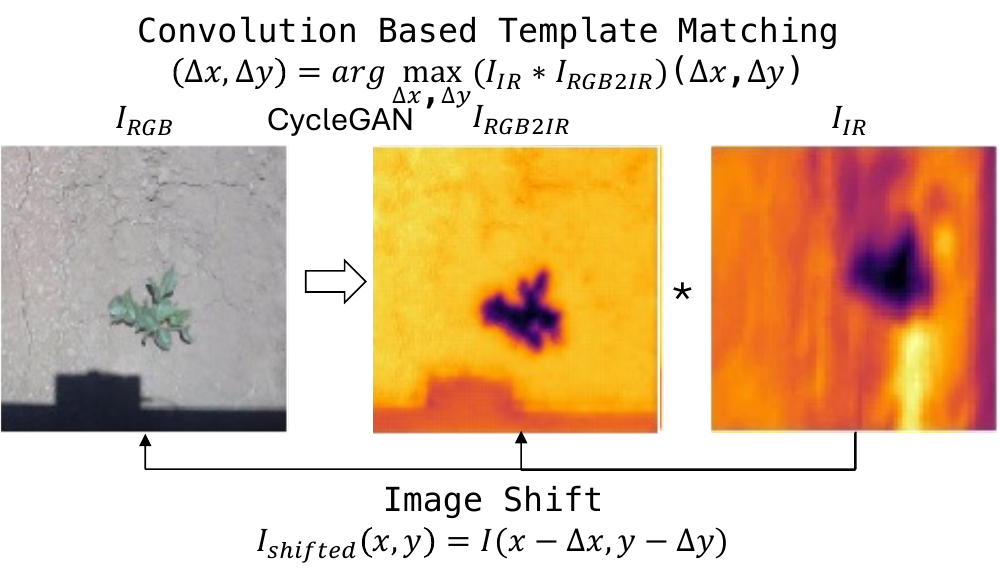}
    \caption{Matching RGB and thermal images using CycleGAN and template matching}
    \label{fig:cycleGanTemplateMatching}
\end{figure}

Therefore, a template matching method based on image intensity was employed to align the domain-transformed image and the thermal image, which yielded more stable results compared to other methods. Figure \ref{fig:cycleGanTemplateMatching} illustrates the input RGB image, the RGB-to-thermal image translated by Cycle GAN, and the low-resolution thermal image to be aligned. Inspired by the approach of Arar et al.~\cite{ararUnsupervisedMultiModalImage2020}, the RGB image was first translated to the thermal imaging camera's domain using Cycle GAN~\cite{zhuUnpairedImageToImageTranslation2017}. Then, template matching was performed between the domain-translated RGB image and the input low-resolution thermal image. The maximum correlation value was calculated based on the image convolution operation from one image to another, which can be hardware-accelerated and integrated into a super-resolution module using PyTorch.

After aligning the domain-transformed RGB image with the thermal image, the original RGB image was also transformed using the alignment result. Subsequently, the RGB, domain-transformed, and low-resolution thermal images were combined and inputted into a ResNet-based Convolutional Neural Network (CNN). The output image was then fed into a Discriminator CNN for Generative Adversarial Network (GAN) training. This architecture is depicted in Figure \ref{fig:front}, referred to as VisTA SR.


Except for CycleGAN~\cite{zhuUnpairedImageToImageTranslation2017} and Template Matching, the implementation followed that of SRGAN~\cite{ledigPhotoRealisticSingleImage2017} and ESRGAN~\cite{wangESRGANEnhancedSuperResolution2018}, and the loss function used is as follows:

Cycle Consistency Loss~\cite{zhuUnpairedImageToImageTranslation2017}:
\begin{equation}
    l_{\text{Consi}}^{\text{Cycle}} = \left|I_{\text{RGB}} - G_{\text{IR2RGB}}(G_{\text{RGB2IR}}(I_{\text{RGB}}))\right|
\end{equation}

Identity Loss~\cite{zhuUnpairedImageToImageTranslation2017}:
\begin{equation}
    l_{\text{MSE}}^{\text{Cycle}} = \left|\left|I_{\text{HR}} - G_{\text{RGB2IR}}(I_{\text{RGB}})\right|\right|
\end{equation}

MSE Loss~\cite{ledigPhotoRealisticSingleImage2017,wangESRGANEnhancedSuperResolution2018}:
\begin{equation}
    l_{\text{MSE}}^{\text{SR}} = \left|\left|I_{\text{HR}} - G_{\text{SR}}(I_{\text{LR}}, I_{\text{RGB}})\right|\right|
\end{equation}

Content Loss (\cite{ledigPhotoRealisticSingleImage2017,wangESRGANEnhancedSuperResolution2018}):
\begin{equation}
    l_{\text{VGG}}^{\text{SR}} = \left|\left|\phi_{\text{VGG}}(I_{\text{HR}}) - \phi_{\text{VGG}}(G_{\text{SR}}(I_{\text{LR}}, I_{\text{RGB}}))\right|\right|
\end{equation}

Adversarial Loss~\cite{ledigPhotoRealisticSingleImage2017,wangESRGANEnhancedSuperResolution2018}):
\begin{equation}
    l_{\text{Adv}}^{\text{SR}} = -\log D_{\text{SR}}(G_{\text{SR}}(I_{\text{LR}}, I_{\text{RGB}}))
\end{equation}

Total Loss:
\begin{equation}
    l_{\text{total}, G} = (l_{\text{Re}}^{\text{Cycle}} + l_{\text{MSE}}^{\text{Cycle}}) + (l_{\text{MSE}}^{\text{SR}} + l_{\text{VGG}}^{\text{SR}} + {\alpha} l_{\text{Adv}}^{\text{SR}})
\end{equation}

\section{Results}
\subsection{Low-Cost Thermal Camera Field Validation with High Fidelity Thermocouple Camera}

\begin{figure}[t]
    \centering
    \includegraphics[width=1.0\linewidth]{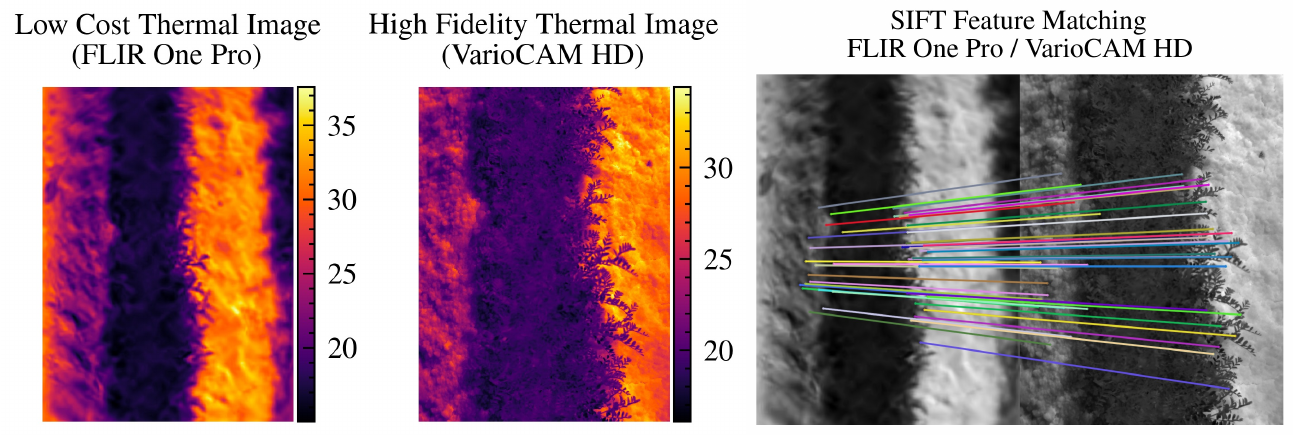}
     \caption{An example of feature matching based temperature comparison between FLIR One Pro and VarioCam HD Camera}
     \label{fig:calibration_field_validation}
\end{figure}

Field data was collected to validate the temperature accuracy of the low-resolution thermal imaging camera in a real-world environment with crops and soil. The data was collected in the Garbanzo bean (\textit{Cicer arietinum}) field located in Davis, California. The ground truth temperature values were measured using a VarioCAM HD camera and compared with the temperature measured by the FLIR One Pro thermal camera, and a total of 170 image pairs were collected on April 5, 2022. Image feature points were extracted from both images using the SIFT~\cite{loweDistinctiveImageFeatures2004} feature extractor, and they were matched using the Flann matching algorithm~\cite{FastApproximateNearest2009}. Then, the homography between the two images was calculated, and outliers were removed using the RANSAC algorithm~\cite{fischlerRandomSampleConsensus1981}. As a result, the temperature values from the corresponding points in the two images were compared (Figure \ref{fig:calibration_field_validation}). 

\begin{figure}
    \centering
    \includegraphics[width=0.8\linewidth]{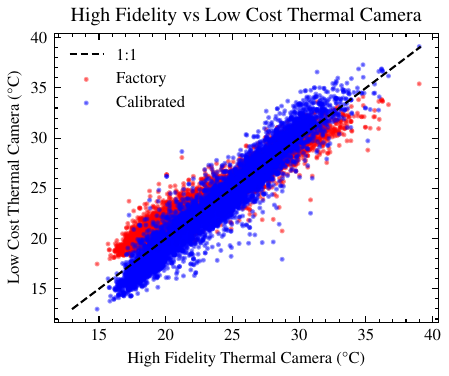}
     \caption{Feature matching based temperature comparison result. Plotted all matched temperature points for a total of 170 images}
     \label{fig:VarioCAMvsFLIROne}
\end{figure}

The matching result for the 170 image pairs is shown in Figure \ref{fig:VarioCAMvsFLIROne}, and Table \ref{tab:calibration_result} summarizes the results. It indicates that the temperature measurement accuracy was improved from $R^2=0.86$ to $R^2=0.89$ after calibration, and the Root Mean Square Error (RMSE) was also improved from \SI{1.52}{\celsius} to \SI{1.40}{\celsius}. Since using the factory parameters tends to overestimate the temperature when it is below \SI{20}{\celsius}, as shown in Figure \ref{fig:calibration_1on1}, the temperature values obtained using the factory parameters in Figure \ref{fig:calibration_field_validation} also showed higher temperature measurements than the actual temperatures. 

Table \ref{tab:calibration_result} also indicates that when calculating RMSE and $R^2$ using only data between \SI{15}{\celsius} and \SI{30}{\celsius}, the temperature measurements with calibrated parameters showed better accuracy. Considering the typical leaf temperature of plants, the accuracy within this temperature range is crucial for thermal cameras used in agriculture. Therefore, the thermal camera calibration in this study demonstrates the potential to enhance temperature measurement accuracy in agricultural research.

\begin{table}
    \centering
    \begin{tabular}{@{}lcccc@{}}
    \toprule
    & \multicolumn{2}{c}{All data} & \multicolumn{2}{c}{\SI{15}{\celsius} - \SI{30}{\celsius}}\\
    & $R^2$ & RMSE (\SI{}{\celsius} )& $R^2$ & RMSE (\SI{}{\celsius})\\
    \midrule
    Factory & 0.86 & 1.52 & 0.83 & 1.52\\   
    Calibrated & 0.89 &1.40 & 0.86 & 1.39\\
    \bottomrule
    \end{tabular}
    \caption{Low-cost thermal camera (FLIR One Pro) temperature accuracy validation result before and after parameter calibration}
    \label{tab:calibration_result}
\end{table}

\subsection{VisTA SR Result}
\begin{figure*}
    \centering
    \includegraphics[width=1.0\linewidth]{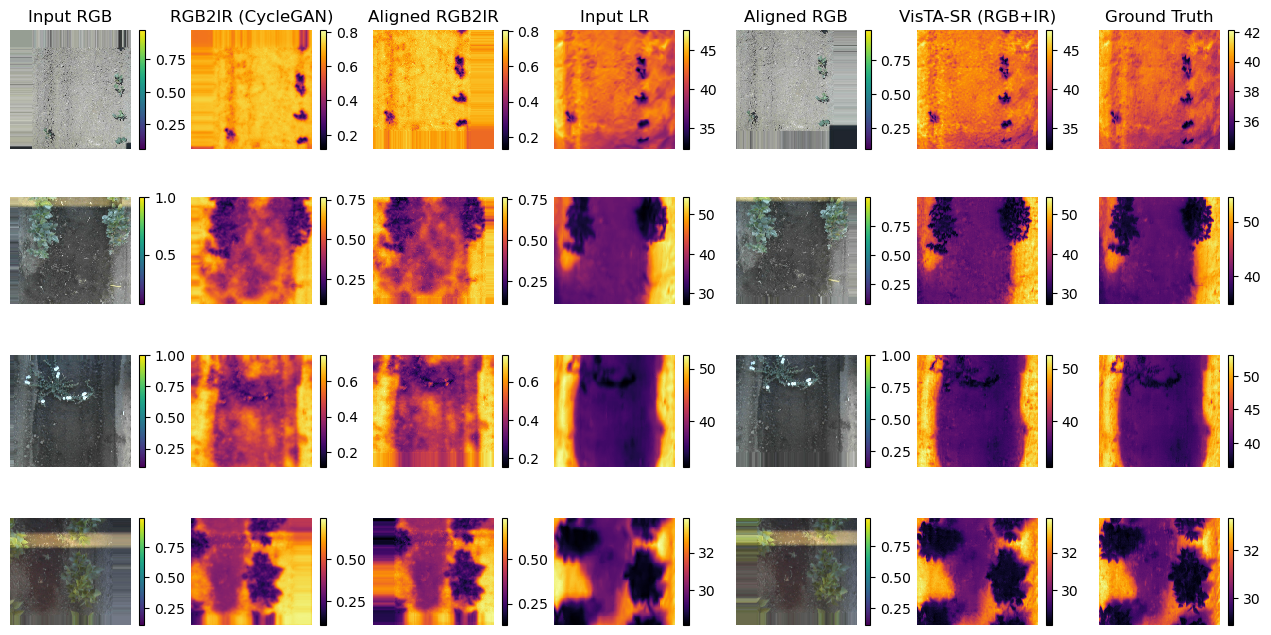}
     \caption{Input low-resolution images, domain translated images, and aligned images using CycleGAN and template matching}
     \label{fig:CycleGAN_TM_Result}
\end{figure*}

In 2022, a total of 2612 image pairs were collected from a warm-season grain legume field across the growing season by matching low-resolution(160x120, FLIR One Pro) thermal images with high-resolution (640X512, FLIR Boson) thermal images. 80\% of these pairs were used as training data, while the remaining 20\% were used for validation. The network was trained over 200 epochs with a batch size of 4. Figure \ref{fig:CycleGAN_TM_Result} demonstrates the image conversion quality and image alignment performance of the CycleGAN module, which was trained simultaneously with the SR Network. As depicted in the example images, CycleGAN successfully translated the image domain and template matching successfully aligned low-resolution thermal images based on image intensity.

\begin{figure}
    \centering
    \includegraphics[width=1.0\linewidth]{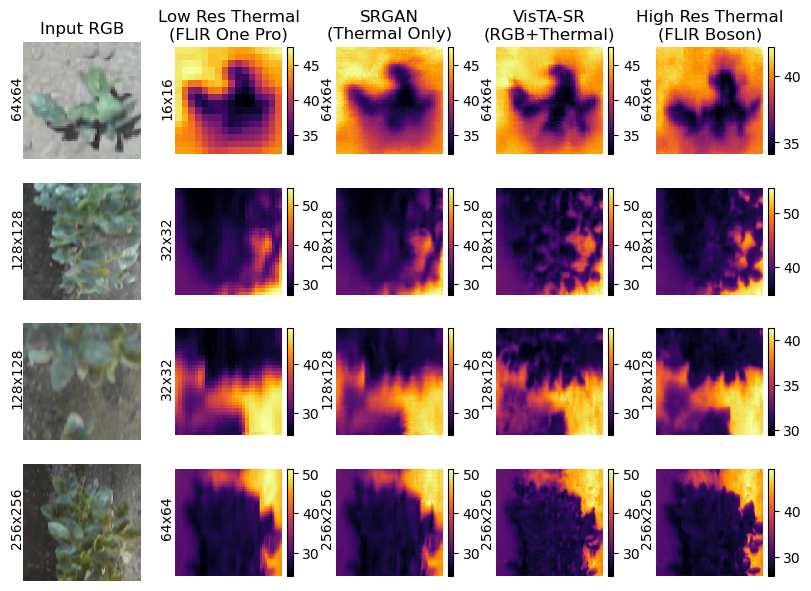}
     \caption{Comparison of input RGB, low-resolution thermal image input, SRGAN\cite{ledigPhotoRealisticSingleImage2017} output in multiple image scales (64x64, 128x128, and 256x256), VisTA-SR output, and ground truth high-resolution thermal image}
     \label{fig:VisTASR_Results2}
\end{figure}

Figure \ref{fig:VisTASR_Results2} presents the results from the multiple input image scales obtained from the VisTA SR algorithm using the input of the combined RGB image aligned with CycleGAN and Template Matching, compared to the results of the Super-Resolution Generative Adversarial Network (SRGAN) algorithm~\cite{ledigPhotoRealisticSingleImage2017, wangESRGANEnhancedSuperResolution2018} that utilizes only the existing thermal image modality. Our VisTA-SR demonstrated higher sharpness by leveraging higher-frequency structural information from the RGB image. This demonstrates that VisTA-SR improved the performance of capturing thermal properties of smaller features at the organ level, as opposed to the plant level.

Table \ref{tab:super-resolution} compares the performance of Bilinear interpolation, Super-Resolution Generative Adversarial Network (SRGAN), and our proposed VisTA SR algorithm, where the Bilinear interpolation method exhibited the highest Root Mean Square Error (RMSE) but the highest Structural Similarity Index (SSIM\cite{wangImageQualityAssessment2004}) and the lowest Peak Signal-to-Noise Ratio (PSNR), while SRGAN and VisTA SR demonstrated similar performance with an RMSE of \SI{2.75}{\celsius}.
It can be inferred that the higher RMSE value of the Bilinear algorithm is because SRGAN and VisTA SR learned the temperature distribution of the training dataset, and Bilinear's higher SSIM value is believed to be a result of the original dataset already being aligned with the template matching process.
Additionally, SRGAN showed a higher PSNR value than VisTA SR, but VisTA SR exhibited excellent visual quality, indicating that evaluating the performance of the Super-Resolution (SR) algorithm solely based on these image metrics is not ideal.

\begin{table}[h]
    \centering
    \begin{tabular}{lrrr}
    \toprule
    Technique & RMSE (\SI{}{\celsius}) & SSIM & PSNR \\
    \midrule
    Bilinear & 2.84 & 0.74 & 23.84 \\
    SRGAN\cite{wangESRGANEnhancedSuperResolution2018} & 2.74 & 0.63 & 24.26 \\
    VisTA SR (Ours) & 2.75 & 0.63 & 23.67 \\
    \bottomrule
    \end{tabular}
    \caption{RMSE, SSIM, and PSNR comparison of Bilinear, SRGAN, and VisTA SR algorithms}
    \label{tab:super-resolution}
\end{table}
\section{Conclusion \& Future Work}
This paper proposes a method to enhance temperature accuracy and image sharpness using a low-resolution thermal imaging camera for agricultural image acquisition. First, we conducted a calibration process to improve the temperature accuracy of the low-resolution thermal imaging camera, followed by field experiments for validation. It is confirmed that the temperature accuracy improved when using the calibrated parameters. We propose the VisTA-SR algorithm for converting low-resolution thermal images to high-resolution ones by aligning and combining RGB and low-resolution images. Through such improvements in temperature accuracy and image sharpness, we will be able to detect small temperature differences between crop tissues or parts, and analyze them in relation to genotypes, growth environments, growth stages, and various other factors.

One limitation was the difficulty of evaluating the performance of super-resolution algorithms in agricultural data using existing image metrics. Since most super-resolution studies generate low-resolution images by down-sampling the high-resolution images. In this case, the pixels of the low-resolution and high-resolution pairs are perfectly aligned, so the image metrics are proportion to the algorithm's super-resolution performance. However, in our study, low-resolution thermal images were actually collected with high-resolution images. Therefore, the output result of our algorithm from low-resolution input may not have a perfect pixel match with the high-resolution image. Considering the characteristics of those image metrics that change significantly even by a few pixel changes, it can be inferred that the image evaluation metrics used in Table \ref{tab:super-resolution} reflected errors derived from multiple camera systems problem, even if the VisTA-SR had excellent visual quality result than others. However, from an agricultural research perspective, temperature accuracy and the ability to detect plants are important for understanding their complex biophysical characteristics. In other words, developing specialized thermal image metrics for agricultural data that reflect these features for performance evaluation in future research is necessary. In future studies, we will examine whether the thermal image improvement algorithm maintains, improves, or hallucinates temperature information in thermal images. Also, using the thermal images processed with the algorithm developed in this paper, we will estimate biophysical parameters such as stomatal conductance in plants and compare accuracy with original and high-resolution image inputs.

\section{Acknowledgement}
This work was financially supported by the Bill and Melinda Gates Foundation, Project ID: INV- 002830, G×E×M Innovation in Intelligence for Climate Adaptation.
{
    \small
    \bibliographystyle{ieeenat_fullname}
    \bibliography{main, Thermal_SR_Paper}
}


\end{document}